# A NOTE ON THE MEASURE OF DISCORD

George J. Klir
*Department of Systems Science*
*Thomas J. Watson School of*
*Engineering and Applied Sciences*
*State University of New York*
*Binghamton, New York 13902, U.S.A.*

Behzad Parviz
*Department of Mathematics*
*and Computer Science*
*California State University*
*Los Angeles, California 90032, U.S.A.*

## Abstract

*A new entropy-like measure as well as a new measure of total uncertainty pertaining to the Dempster-Shafer theory are introduced. It is argued that these measures are better justified than any of the previously proposed candidates.*

This is a short note on the generalization of the Shannon entropy from probability theory to the **Dempster-Shafer theory (DST)**. We assume that the reader is familiar with the fundamentals of these theories (Klir and Folger 1988).

It is now well known that two types of uncertainty coexist in DST, which are usually referred to as **nonspecificity** (or imprecision) and **conflict** (or discord, dissonance). Their measures are overviewed in a recent paper (Klir and Ramer 1990), which is an essential companion to this note.

It is well established that nonspecificity in DST is properly measured by a function N defined by the formula

$$N(m) = \sum_{A \in F} m(A) \log_2 |A|, \quad (1)$$

where m, F, |A| denote, respectively, a given basic probability assignment function in DST, a set of focal subsets induced by m, and the number of elements (cardinality) of a focal subset A; it is assumed that m is defined on the power set of a finite universal set X.

Function N was proven unique under appropriate requirements; it is additive, subadditive, monotonic (Dubois and Prade 1987), and its range is

$$0 \leq N(m) \leq \log_2 |X|. \quad (2)$$

In addition, N measures nonspecificity in convenient units (bits), and it is a natural generalization of the Hartley measure of uncertainty (Klir and Folger 1988) from classical set theory to random set theory, which forms a base of DST.

The question of how to measure the second type of uncertainty, which is connected with conflicts among evidential claims, has been far more controversial. Although it is generally agreed that a measure of this type of uncertainty must be a generalization of the well-established Shannon entropy (Klir and Folger 1988) from probability theory to DST, the question is which of the proposed measures, which all collapse to the Shannon entropy within the domain of probability theory, is the right generalization.

Two distinct measures, both seemingly generalizations of the Shannon entropy in DST, were prepared in the early 1980s (Klir and Folger 1988, Klir and Ramer 1990, Dubois and Prade 1987). As argued in the companion paper (Klir and Ramer 1990), either of these measures is deficient in some traits. To alleviate these deficiencies, another measure was prepared in the companion paper, which is called a measure of **discord**. This measure is expressed by a function D defined by the formula

$$D(m) = -\sum_{A \in F} m(A) \log_2 \left[ 1 - \sum_{B \in F} m(B) \frac{|B-A|}{|B|} \right]. \quad (3)$$

The rationale for choosing this function is explained as follows. The term

$$Con(A) = \sum_{B \in F} m(B) \frac{|B-A|}{|B|} \quad (4)$$

in Eq. (3) expresses the sum of conflicts of individual evidential claim m(B) for all B ∈ F with respect to the evidential claim m(A) focusing on a particular set A; each individual conflict is properly scaled by the degree to which the subsethood B ≤ A is violated. The function

$$-\log_2[1 - Con(A)],$$

which is employed in Eq. (3), is monotonic increasing with Con(A) and, consequently, it represents the same quantity as Con(A), but on the logarithmic scale. The use of the logarithmic scale is motivated in the same way as in the case of the Shannon entropy (Klir and Folger 1988). Indeed, the Shannon entropy, which is applicable only to functions m defined on singletons, can be expressed in the form comparable with Eq. (3):

$$D(m) = -\sum_{x \in X} m(\{x\}) \log_2 [1 - \sum_{y \neq x} m(\{y\})]. \quad (5)$$

Function D is clearly a measure of the average conflict among evidential claims within a given body of evidence. The function can also be expressed in a simpler form,

$$D(m) = -\sum m(A) \log_2 \sum m(B) \frac{|A \cap B|}{|B|}, \quad (6)$$

which follows immediately from Eq. (3).

In addition to its intuitive appeal as a measure of average conflict, function D possesses some desirable mathematical properties (Klir and Ramer 1990, Ramer and Klir 1992): it is additive, its range is $[0, \log_2 |X|]$, its measurement units are bits and, as already mentioned, it is equivalent to the Shannon entropy within the restricted domain of probability theory. In possibility theory, D(m) is bounded from above: it converges to a constant, estimated as 0.892, as $|X| \to \infty$ (Geer and Klir 1991).

Considering the two types of uncertainty in DST, non-specificity and conflict, it is natural to express the total uncertainty in DST, T(m), as the sum

$$T(m) = N(m) + D(m). \quad (7)$$

Function T is additive, its range is (surprisingly) the same as the range of its components (i.e., $[0, \log_2 |X|]$), and its measurement units are again bits (Ramer and Klir 1992, Ramer 1991).

Although functions D and T are improvements of their previously considered counterparts (Klir and Ramer 1990, Ramer and Klir 1992), they still have one deficiency: neither of them is subadditive and it is relatively easy to generate examples in which subadditivity is violated. While it may be argued that subadditivity is not essential in this case, we have not been able to find any way to justify this argument. A similar sentiment is expressed in a paper by Vejnarova (Vejnarova 1991), who shows explicitly that function N is subadditive, but functions D and T are not.

This lack of complete satisfaction with functions D and T, reinforced by Vejnarova (Vejnarova 1991), led us to a further reexamination of the notion of entropy-like measure in DST. As a result, we found the following conceptual defect in function D as a measure of conflict.

Let sets A and B in Eq. (4) be such that A ⊂ B. Then, according to function Con, the claim m(B) is taken to be in conflict with the claim m(A) to the degree $|B-A|/|B|$. This, however, should not be the case: the claim focusing on B is implied by the claim focusing on A (since A ⊂ B) and, hence, m(B) should not be viewed in this case as contributing to the conflict with m(A).

Consider, as an example, incomplete information regarding the age of a person, say Joe. Assume that the information is expressed by two evidential claims pertaining to the age of Joe: "Joe is between 15 and 17 years old" with degree m(A), where A = [15, 17], and "Joe is a teenager" with degree m(B), where B = [13, 19]. Clearly, the weaker second claim does not conflict with the stronger first claim.

Assume now that A ⊃ B. In this case, the situation is inverted: the claim focusing on B is not implied by the claim focusing on A and, consequently, m(B) does conflict with m(A) to a degree proportional to number of elements in A that are not covered by B. This conflict is not captured by function Con since $|B-A| = 0$ in this case.

It follows from these observations that the total conflict of evidential claims within a body of evidence (F, m) with respect to a particular claim m(A) should be expressed by function

$$CON(A) = \sum_{B \in F} m(B) \frac{|A - B|}{|A|} \quad (8)$$

rather than function Con given by Eq. (4). Replacing Con(A) in Eq. (3) with CON(A), we obtain a new function, which is better justified as a measure of conflict in DST than function D. This new function, which we suggest to call strife and denote by S, is defined by the form

$$S(m) = -\sum_{A \in F} m(A) \log_2 \left[ 1 - \sum_{B \in F} m(B) \frac{|A - B|}{|A|} \right]. \quad (9)$$

It is trivial to convert this form into a simpler one,

$$S(m) = -\sum_{A \in F} m(A) \log_2 \sum_{B \in F} m(B) \frac{|A \cap B|}{|A|}, \quad (10)$$

where the term $|A \cap B|/|A|$ expresses the degree of subsethood of set A in set B. Eq. (10) can also be rewritten as

$$S(m) = N(m) - \sum_{A \in F} m(A) \log_2 \sum_{B \in F} m(B) |A \cap B|, \quad (11)$$

where N(m) is the nonspecificity measure given by Eq. (1). Furthermore, introducing

$$K(m) = \sum_{A \in F} m(A) \log_2 \sum_{B \in F} m(B) |A \cap B|, \quad (12)$$

we have

$$S(m) = N(m) - K(m).$$

Let the total uncertainty in DST based upon nonspecificity N and strife S be denoted by NS. Then,

$$NS(m) = N(m) + S(m)$$

or, alternatively,

$$NS(m) = 2N(m) - K(m).$$

Substituting for N(m) and K(m) from Eqs. (1) and (12), we obtain

$$NS(m) = \sum_{A \in F} m(A) \log_2 \sum_{B \in F} m(B) \frac{|A|^2}{|A \cap B|}, \quad (13)$$

It is reasonable to conclude that functions S and NS are well justified on intuitive grounds. The question is whether they also possess essential mathematical properties. The following propositions (given here without proofs) and conjectures (supported by ample evidence) give at least a partial answer to this question:

1. It is easily verifiable that the measurement units of both S and NS are bits.

2. Whenever m defines a probability measure (i.e., all focal subsets are singletons), both S and NS assume the form of the Shannon entropy.

3. The range of S is $[0, \log_2 |X|]$. A proof of this proposition is similar to the proof of the analogous proposition for function D in (Ramer and Klir 1992). $S(m) = 0$ iff $m(A) = 1$ for some $A \subseteq X$; $S(m) = \log_2 |X|$ iff $m(\{x\}) = 1/|X|$ for all $x \in X$.

4. The range of NS is $[0, \log_2 |X|]$. Although this proposition is only a conjecture at this time, we expect that it can be proven in a similar way as the analogous proposition for T in (Ramer 1991). The minimum, $NS(m) = 0$, is obtained iff $m(\{x\}) = 1$ for some $x \in X$, which is the correct representation of complete certainty (full information). The maximum, $NS(m) = \log_2 |X|$, is obtained for all bodies of evidence (m, F) such that:

(a) m is uniformly distributed among all focal subsets (elements of F); and

(b) F is strongly symmetric in the following sense:

(b.1)   All focal subsets have the same cardinality;
(b.2)   each element of X belongs to the same number of focal subsets.

Examples of families of focal subsets F that satisfy this strong symmetry are:

- any partition of X into blocks of equal cardinality;
- for each $k = 1,2,\ldots, |X|$, the family of all subsets with cardinality k;
- for each $k = 1,2,\ldots,n$, the family $\{\{x_{1+j(\text{mod } n)}, x_{2+j(\text{mod } n)},\ldots, x_{k+j(\text{mod } n)}\} \mid j = 0, 1, \ldots, n-1\}$, where $X = \{x_1, x_2,\ldots,x_n\}$ (let this family be called a chain of subsets of cardinality k);
- any partition of X into blocks of equal cardinality c, where each block contributes to F all its subsets with cardinality k ($k = 1,2,\ldots c$);
- any partition of X into blocks of equal cardinality c, where each block contributes to F the chain of subsets of a particular cardinality k ($k = 1,2,\ldots,c$).

All these examples conform perfectly to our intuitive perception of maximum uncertainty in DST. Whether they cover all bodies of evidence for which $NS(m) = \log_2 |X|$ has yet to be determined, but it is quite likely that they do.

5. Both S and NS are additive. The additivity of N is well established (Klir and Folger 1988, Dubois and Prade 1987) and the additivity of S can be proven in a way analogous to the proof of additivity of D in (Klir and Ramer 1990). Additivity of NS follows from the additivity of N and S.

6. Counterexamples demonstrating that S is not subadditive are relatively easy to find. Consider, for example, the following joint basic assignment function defined on $X \times Y$, were $X = \{a, b\}$ and $Y = \{\alpha, \beta\}$: $m(X \times Y) = 0.5$ and $m(\{a, \alpha\}, \{b, \beta\}) = 0.5$. Then, $S(m) = 0.5 (\log_2 4 - \log_2 3) > 0$. However, $m_x(X) = 1$ and $m_y(Y) = 1$ and, consequently, $S(m_x) + S(m_y) = 0$ and Klir and Ramer, 1990 violated. Although Arthur Ramer constructed some counterexamples demonstrating that neither NS is subadditive (personal communication) these examples seem to be rather rare and the violation of subadditivity is very small relative to the amounts of uncertainty involved.

7. Employing ordered possibility distributions $1 = r_1 \geq r_2 \geq \ldots \geq r_n$, the forms of S and NS in possibility theory are very simple:

$$S(m) = N(m) - \sum_{i=2}^{n} (r_i - r_{i+1}) \log_2 \sum_{j=1}^{i} r_j, \quad (14)$$

$$NS(m) = 2N(m) - \sum_{i=2}^{n} (r_i - r_{i+1}) \log_2 \sum_{j=1}^{i} r_j, \quad (15)$$

where $N(m)$ is the measure of nonspecificity, which in possibility theory has the form

$$N(m) = \sum_{i=2}^{n} (r_i - r_{i+1}) \log_2 i \quad (16)$$

($r_{n+1} = 0$ by convention in Eqs. (13) - (15)).

8. The maximum value of possibilistic strife, given by Eq. (14), depends on n in the same way as the maximum value of possibilistic discord (Geer and Klir 1991): it increases with n and converges to a constant, estimated as 0.892, as $n \to \infty$. However, the possibility distributions for which the maxima of possibilistic strife are obtained (one for each value of n) are different from those for possibilistic discord.

These properties and the intuitive justification of functions S and NS make these functions better candidates for the entropy-like measure and the measure of total uncertainty in DST than any of the previously considered functions (Klir and Folger 1988, Klir and Ramer 1990, Dubois and Prade 1987, Ramer and Klir 1992).